\documentclass{article}

\usepackage{paper}
\usepackage[utf8]{inputenc}
\usepackage[T1]{fontenc}

\usepackage{amsmath,amsfonts,bm}

\def\eqref#1{equation~\ref{#1}}

\def\1{\bm{1}}

\DeclareMathAlphabet{\mathsfit}{\encodingdefault}{\sfdefault}{m}{sl}
\SetMathAlphabet{\mathsfit}{bold}{\encodingdefault}{\sfdefault}{bx}{n}

\usepackage{graphicx}
\usepackage{booktabs}
\usepackage{amssymb}
\usepackage{multirow}
\usepackage{microtype}
\usepackage[table,dvipsnames]{xcolor}
\usepackage{xspace}
\usepackage{url}
\usepackage{fontawesome}
\definecolor{linkblue}{rgb}{0.09,0.5,0.99}
\usepackage[breaklinks,colorlinks,allcolors=linkblue]{hyperref}

\usepackage[font=small]{caption}
\usepackage{enumitem}
\setlist[itemize,1]{leftmargin=\dimexpr 18pt}
\setlist[enumerate,1]{leftmargin=\dimexpr 18pt}

\usepackage[
  backend=biber,
  citestyle=numeric,
  bibstyle=numeric,
  mincitenames=1,
  maxcitenames=1,
  maxbibnames=99,
]{biblatex}
\newcommand{\citep}[1]{\parencite{#1}}

\definecolor{MyRed}{HTML}{D2042D}
\definecolor{MyGreen}{HTML}{00A36C}
\definecolor{MyBlue}{HTML}{1F51FF}
\definecolor{MyPurple}{HTML}{AF58BA}
\definecolor{MyYellow}{HTML}{FFC61E}
\definecolor{MyOrange}{HTML}{F28522}
\definecolor{MyGrey}{HTML}{A0B1BA}
\definecolor{MyBrown}{HTML}{A6761D}
\definecolor{lightgrey}{rgb}{0.83, 0.83, 0.83}
\definecolor{lavendermist}{rgb}{0.9, 0.9, 0.98}
\definecolor{lightblue}{HTML}{dfebf7}
\definecolor{lightgreen}{HTML}{f8fcf4}
\newcommand{\red}[1]{{\color{MyRed}#1}}

\newcommand{\purple}[1]{{\color{MyPurple}#1}}

\newcommand{\grey}[1]{{\color{MyGrey}#1}}

\newcommand{\na}{\grey{N/A}}

\newcommand{\model}{\textsc{\texttt{HuGe}}\xspace}

\newcommand{\projectlinks}[2]{%
  \faHome\,\,\href{#1}{Project Page}\quad
  \faGithub\,\,\href{#2}{Code}%
}

\graphicspath{{figures/}}

\title{\model: Scaling Humanoid Generalist Vision-Language-Action Models with Egocentric Experience}

\author{
\normalfont Yang Liu$^{1}$\thanks{Equal contribution.\quad\textsuperscript{\ensuremath{\dagger}}Corresponding authors.} \, Jiacheng Liu$^{1,2,3}$\footnotemark[1] \, Hengtao Li$^{4}$\footnotemark[1] \, Yuheng Huang$^{2}$ \, Zijian Yang$^{5}$ \, Ji Dai$^{5}$ \\[1mm]
Maoxin Wang$^{5}$ \, Yi Gao$^{6}$ \, Qishen Wang$^{7}$ \, Baohua Yin$^{8}$ \, Peiyang Li$^{5}$ \, Shuanghao Bai$^{9}$ \\[1mm]
Mingyang Sun$^{1,2,3}$ \, Pengxiang Ding$^{1,2}$ \, Jianqin Yin$^{5}$ \, Lixin Yang$^{3,10}$\footnotemark[2] \, Donglin Wang$^{1}$\footnotemark[2] \\[1mm]
\small $^{1}$Westlake University \quad $^{2}$Zhejiang University \quad $^{3}$Shanghai Innovation Institute \\[1mm]
\small $^{4}$Hebei University of Technology \quad $^{5}$Beijing University of Posts and Telecommunications \\[1mm]
\small $^{6}$Southwest University of Finance and Economics \quad $^{7}$University of Science and Technology of China \\[1mm]
\small $^{8}$University of Sussex \quad $^{9}$Xi'an Jiaotong University \quad $^{10}$Shanghai Jiao Tong University \\[1mm]
\small liuyang67@westlake.edu.cn \quad siriusyang@sjtu.edu.cn \quad wangdonglin@westlake.edu.cn \\[1mm]
\small \projectlinks{https://yliu-cs.github.io/HuGe}{https://github.com/yliu-cs/HuGe}
}

\date{}
\renewcommand{\headeright}{\includegraphics[height=1.2em]{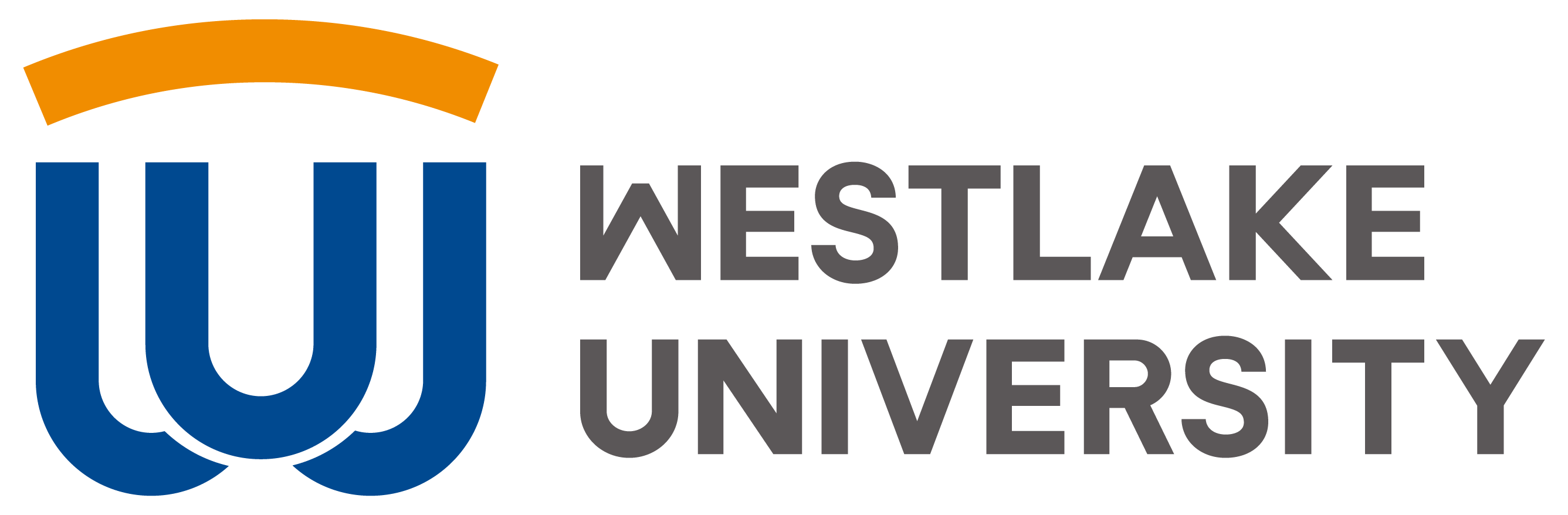}}
\renewcommand{\undertitle}{}
\renewcommand{\shorttitle}{HuGe}

\begin{document}

\maketitle

\begin{abstract}

Extending Vision-Language-Action models (VLAs) to humanoid robots requires coordinated whole-body control and diverse demonstrations that are costly to collect in real-world scenarios.
Although egocentric human demonstrations provide a complementary source of task experience, differences in visual embodiment and missing robot proprioception complicate transfer to humanoid policies.
To address these challenges, we introduce \model, a family of \textbf{Hu}manoid \textbf{Ge}neralist VLAs pre-trained on over 2,000 hours of egocentric human and robot data.
Specifically, \model employs three existing representative VLA architectures to predict latent motion tokens for a pretrained whole-body controller, together with hand-control signals.
Based on this, we introduce a three-stage training pipeline that combines egocentric pre-training, humanoid adaptation, and downstream fine-tuning.
To facilitate transfer across embodiments, we construct pseudo-proprioceptive states through simulation playback during pre-training and modify arm appearance in egocentric observations through re-rendering during downstream fine-tuning.
With this pipeline, \model achieves the state-of-the-art on simulation benchmark HumanoidArena.
Furthermore, real-world experiments on the Unitree G1 indicate that augmenting robot demonstrations with visually adapted egocentric demonstrations significantly increases success on a pick-and-drop task.
\end{abstract}

\section{Introduction}
\label{sec:introduction}

Recent advances in Vision-Language-Action models (VLAs) \citep{DBLP:conf/corl/ZitkovichYXXXXW23,DBLP:conf/corl/KimPKXB0RFSVKBT24,DBLP:journals/corr/abs-2410-24164,DBLP:journals/corr/abs-2503-14734,DBLP:journals/corr/abs-2602-15922,DBLP:journals/corr/abs-2601-21998} fueled by Vision-Language Models (VLMs) \citep{DBLP:journals/corr/abs-2407-07726,DBLP:journals/corr/abs-2511-21631} and World Models (WMs) \citep{DBLP:journals/corr/abs-2503-20314,DBLP:journals/corr/abs-2506-09985} have enabled language-conditioned manipulation across a wide range of tabletop tasks.
Building on this progress, extending VLAs to humanoid robots offers the potential to perform diverse tasks in human environments by combining manipulation with whole-body mobility.
Despite this potential, adapting VLAs to humanoid robots remains challenging because task execution requires coordinated whole-body motion while maintaining balance and stability.
To meet these control requirements, existing humanoid loco-manipulation approaches commonly employ hierarchical architectures that combine high-level policies with reinforcement learning (RL)-based controllers \citep{jiang2026wholebodyvla,DBLP:journals/corr/abs-2603-12263,DBLP:journals/corr/abs-2604-07993}.
To this end, a central research question is how to train generalist VLAs that translate language instructions into coordinated whole-body behaviors across diverse tasks.

Realizing this goal requires addressing two key challenges.
First, the high-dimensional action spaces and complex dynamics of humanoid robots make it difficult to directly transfer VLA policies developed for robotic manipulators to whole-body control.
In particular, arm movements must be coordinated with the torso and lower body to accomplish manipulation objectives while maintaining balance and stability.
Consequently, VLA-based humanoid control requires an action interface that connects task-level predictions to coordinated whole-body execution.
Second, beyond these control challenges, learning generalist humanoid behaviors requires diverse whole-body demonstrations, yet collecting such data on real-world robots is costly and time-consuming.
To alleviate this data bottleneck, egocentric human demonstrations offer a complementary source of task experience for large-scale training.
However, differences in visual embodiment and the absence of robot proprioceptive observations complicate the direct use of these demonstrations for humanoid policy learning.
These challenges motivate an approach that integrates VLA policies with whole-body control and supports effective transfer from egocentric human experience to humanoid robots through a scalable training pipeline.

To address these challenges, we introduce \model, a family of \textbf{Hu}manoid \textbf{Ge}neralist VLAs that operate through a whole-body controller and learn from a large-scale mixture of egocentric human and robot data.
Specifically, \model builds on existing representative VLA architectures and follows a three-stage training pipeline that sequentially incorporates egocentric human data, humanoid teleoperation data, and a mixture of downstream demonstrations.
To support this pipeline, we assemble a training corpus comprising over 2,000 hours of egocentric and robot data.
Furthermore, we prepare egocentric data for humanoid policy learning during both pre-training and downstream fine-tuning.
In particular, during pre-training, we use simulation playback to construct pseudo-proprioceptive states for egocentric data that lack robot proprioceptive observations.
Building on this pre-training stage, we re-render and replace arm regions in egocentric observations during downstream fine-tuning.
To assess the effectiveness of this approach, we compare \model with baseline methods on the HumanoidArena simulation benchmark and further evaluate its performance on real-world humanoid tasks.
Overall, \model achieves the state-of-the-art performance on HumanoidArena, demonstrating its capability for humanoid whole-body control, while comprehensive real-world experiments and solid analyses further demonstrate the effectiveness of the training recipe.

Our main contributions are summarized as follows: \\
$\bullet$ We introduce \model, a family of humanoid generalist VLAs, trained on over 2,000 hours of egocentric human and robot data through a three-stage training pipeline. \\
$\bullet$ We develop data adaptation methods that facilitate the transfer of egocentric human experience to humanoid policy learning. \\
$\bullet$ \model achieves the state-of-the-art performance against baselines on HumanoidArena, while comprehensive real-world experiments and analyses validate our design choices and training paradigm.
\section{Related Work}
\label{sec:related-work}

\subsection{Vision-Language-Action Models}
\label{sec:vla}

Vision-Language-Action models (VLAs) \citep{DBLP:conf/corl/ZitkovichYXXXXW23,DBLP:conf/corl/KimPKXB0RFSVKBT24,DBLP:journals/corr/abs-2410-24164,DBLP:journals/corr/abs-2503-14734} built on Vision-Language Models (VLMs) \citep{DBLP:journals/corr/abs-2407-07726,DBLP:journals/corr/abs-2511-21631} and World-Action Models (WAMs) \citep{DBLP:journals/corr/abs-2602-15922,DBLP:journals/corr/abs-2601-21998,DBLP:journals/corr/abs-2603-16666} based on video-prediction World Models (WMs) \citep{DBLP:journals/corr/abs-2503-20314,DBLP:journals/corr/abs-2506-09985} have advanced rapidly in fixed-base single-arm and bimanual manipulation.
Beyond language-conditioned task execution, these approaches combine robot demonstrations with pretrained multimodal knowledge or predictions of future observations to support policy learning across diverse manipulation tasks \citep{DBLP:conf/iclr/TianYZWL0P25,wang2026unified,DBLP:journals/corr/abs-2603-25406}.
These advances have enabled robotic manipulators to autonomously fold and organize deformable objects such as clothes and towels, perform precise operations such as connector insertion and small-part assembly, and complete long-horizon tasks such as table setting and coffee preparation \citep{DBLP:journals/corr/abs-2511-14759,Lin_2026_CVPR,DBLP:journals/corr/abs-2604-15483,DBLP:journals/corr/abs-2605-10903}.
For humanoid whole-body control, we adopt three representative VLA architectures, OFT \citep{DBLP:journals/corr/abs-2502-19645}, GR00T \citep {DBLP:journals/corr/abs-2503-14734}, and $\pi$ \citep{DBLP:journals/corr/abs-2410-24164}, without introducing additional modules or modifying the architectural designs.
Using these architectures, we investigate humanoid whole-body control and training strategies centered on egocentric human demonstrations, complemented by robot data for embodiment adaptation and downstream task learning.

\subsection{Humanoid Whole-Body Control}
\label{sec:humanoid}

Early reinforcement-learning approaches to humanoid control focused on locomotion, training policies in simulation to maintain balance, follow velocity commands, and adapt to terrain variations during real-world deployment \citep{DBLP:journals/scirobotics/RadosavovicXZDMS24}.
Subsequent work expanded these capabilities to whole-body motion tracking, using retargeted human motions to train controllers that support coordinated teleoperation and autonomous skill learning from demonstrations \citep{DBLP:conf/corl/He0HXZ0KLS24,DBLP:journals/corr/abs-2502-14795,DBLP:journals/scirobotics/LuoYWLCCCLMBPSWDDHSLJHX26,DBLP:journals/corr/abs-2606-03985}.
Building on these control capabilities, one line of hierarchical VLA approaches predicts upper-body manipulation targets alongside lower-body motion commands, relying on reinforcement-learning controllers to execute locomotion and posture adjustments \citep{DBLP:journals/corr/abs-2509-11839,jiang2026wholebodyvla,DBLP:journals/corr/abs-2603-12263,DBLP:journals/corr/abs-2604-07993}.
Alongside this direction, advances in whole-body motion models have supported VLA and world-action approaches that generate full-body motion references or unified motion latents, which whole-body controllers translate into coordinated joint commands \citep{DBLP:journals/corr/abs-2606-09215,DBLP:journals/corr/abs-2608-06375}.
In contrast, we focus on training strategies centered on egocentric human demonstrations for humanoid whole-body control, examining three established VLA architectures without adding modules or modifying the architectural designs.

\subsection{Egocentric Learning}
\label{sec:ego}

Egocentric human demonstrations provide diverse observations and a complementary source of experience for robot learning.
Early approaches use human videos to learn visual representations and motion priors that support downstream policy learning \citep{DBLP:conf/corl/NairRKF022,DBLP:conf/corl/RadosavovicXJAM22,DBLP:conf/iclr/MaSJBK023,DBLP:conf/corl/WangFSZ0XZA23}.
Building on these efforts, more direct approaches learn action prediction from annotated human motions or inferred latent actions, incorporating human and robot data through pre-training, co-training, and task-specific adaptation \citep{DBLP:journals/corr/abs-2507-15597,DBLP:journals/corr/abs-2510-21571,DBLP:journals/corr/abs-2601-12993,DBLP:journals/corr/abs-2602-16710}.
To address embodiment differences, another line of work converts human demonstrations into robot-compatible data through motion retargeting and visual adaptation, including human-arm removal, robot rendering, and embodiment masking \citep{lepert2025phantom,DBLP:conf/icra/KareerPPMCWHX25,DBLP:journals/corr/abs-2602-11464}.
Beyond fixed-base manipulation, related efforts extend human-data learning to humanoid manipulation and loco-manipulation through aligned action representations and staged training with robot demonstrations \citep{DBLP:journals/corr/abs-2603-12263}.
Within this setting, we investigate how egocentric human demonstrations can support humanoid whole-body policy learning through controller-compatible action supervision, simulation-derived pseudo-proprioceptive states, and visual adaptation across training stages.
\section{Methodology}
\label{sec:methodology}

\begin{figure}[!t]
  \centering
  \includegraphics[width=\linewidth]{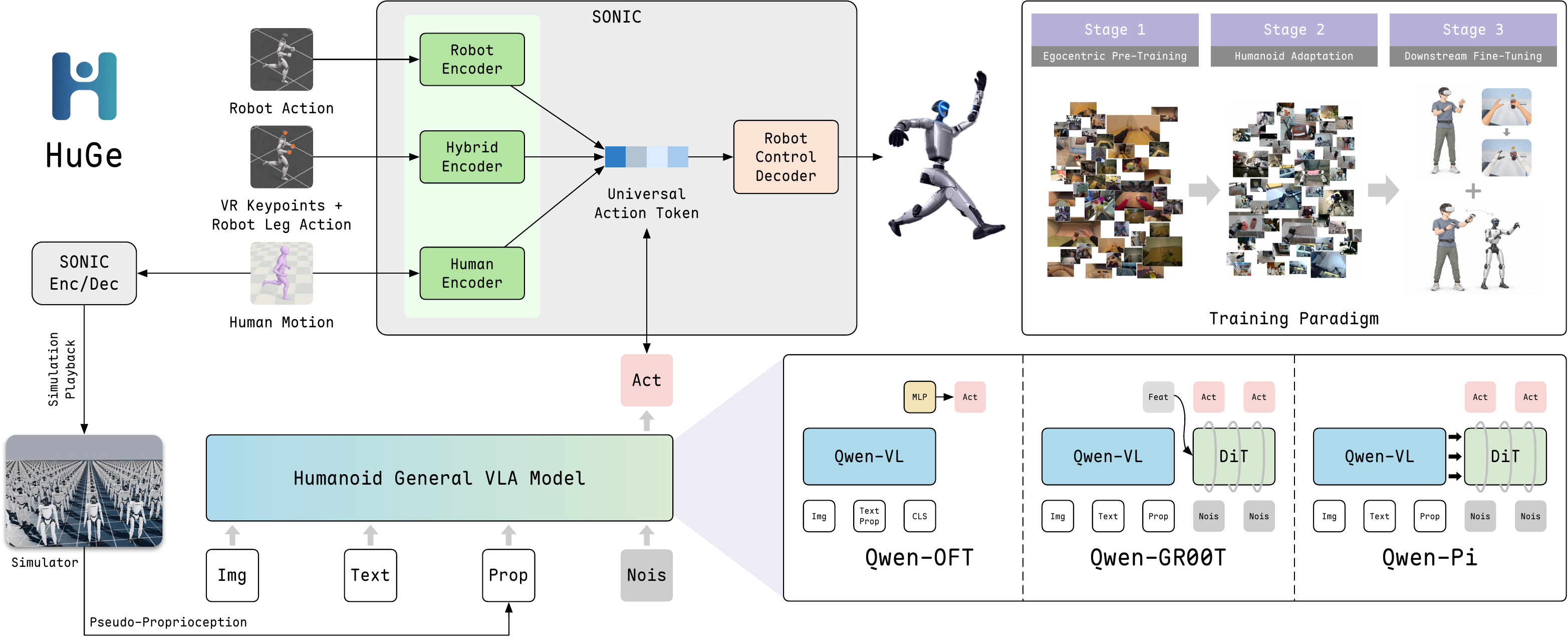}
  \caption{Schematic of the \model framework, architecture and training pipeline.}
  \label{fig:arch}
\end{figure}

In this section, we present the model architecture and training recipe of \model, as shown in Figure~\ref{fig:arch}.
We first describe how three representative VLA architectures are instantiated for humanoid control.
Building on these model instantiations, we introduce a three-stage training pipeline that progressively incorporates egocentric human experience, humanoid demonstrations, and task-specific mixture data.

\subsection{Architecture}
\label{sec:architecture}

Given a visual observation $o_t$, a proprioceptive state $s_t$, and a natural-language task instruction $l$, \model predicts a 66-dimensional action vector $a_t$ for humanoid control at time step $t$.
The action consists of a 64-dimensional latent motion token and two binary variables specifying whether the left and right hands are open or closed.
The motion token lies in the universal token space of SONIC \citep{DBLP:journals/scirobotics/LuoYWLCCCLMBPSWDDHSLJHX26}, a pretrained whole-body controller.
Conditioned on the proprioceptive state of robot, control decoder of SONIC converts the predicted motion token into joint-level commands for whole-body motion, while the two binary variables separately control hand opening and closing.
This interface allows \model to predict task-conditioned whole-body actions while relying on the low-level controller for coordinated motion execution.

We instantiate \model with three existing representative VLA designs: OFT \citep{DBLP:journals/corr/abs-2502-19645}, GR00T \citep {DBLP:journals/corr/abs-2503-14734}, and $\pi$ \citep{DBLP:journals/corr/abs-2410-24164}.
Across these variants, Qwen-based vision-language backbones encode visual observations, proprioceptive state and language instructions, while different action prediction modules map the resulting representations to humanoid actions \citep{DBLP:journals/corr/abs-2604-05014}.
Specifically, the OFT-style variant uses a lightweight multilayer perceptron (MLP) to map the hidden states of predefined action tokens directly to continuous actions, predicting an action chunk in parallel.
In contrast, the GR00T-style variant separates vision-language processing from action generation: the backbone provides contextual features, and a diffusion transformer (DiT) conditions on these features through cross-attention to generate action chunks via flow matching.
Compared with this use of backbone output features, the $\pi$-style variant connects a DiT action expert to hidden states from multiple backbone layers through layer-wise cross-attention, progressively refining noisy action sequences into continuous action chunks.
For humanoid control, all three variants use the action interface defined above, with each predicted time step containing a 64-dimensional latent motion token for SONIC and two hand-control signals.

\subsection{Training Recipe}
\label{sec:training-recipe}

We train \model through a three-stage pipeline that progressively incorporates egocentric human experience, humanoid demonstrations, and task-specific mixture data.
Specifically, egocentric pre-training utilizes large-scale egocentric demonstrations to learn language-conditioned action prediction.
Subsequently, during humanoid adaptation, we continue training the policy on humanoid teleoperation demonstrations to adapt action prediction to robot observations and states.
Finally, downstream fine-tuning adapts the policy to target tasks using task-specific demonstrations.
Table~\ref{tab:training-data} summarizes the data sources and dataset sizes for each training stage.

\begin{table}[!htbp]
  \caption{Training data used across the three-stage training pipeline.}
  \label{tab:training-data}
  \begin{center}
  \small
  \begin{tabular}{@{}llrrr@{}}
    \toprule
    Training Stage & Source & Hours & Episodes & Frames \\
    \midrule
    \multirow{5}{*}{Stage 1}
      & EgoDex \citep{hoque2026egodex}             & 821.800 & 334,991 & 147,923,137 \\
      & EgoVerse \citep{DBLP:journals/corr/abs-2604-07607}            & 47.080  & 1,704   & 5,127,761 \\
      & Xperience-10M \citep{xperience10m}       & 478.330 & 10,651  & 34,589,556 \\
      & BEHAVIOR-1K \citep{li2022behaviork}         & 322.400 & 3,426   & 34,818,979 \\
    \cmidrule(l){2-5}
      & \cellcolor[rgb]{ .949,  .949,  .949}\textsc{Total}
      & \cellcolor[rgb]{ .949,  .949,  .949}\textit{1,669.610}
      & \cellcolor[rgb]{ .949,  .949,  .949}\textit{350,772}
      & \cellcolor[rgb]{ .949,  .949,  .949}\textit{222,459,433} \\
    \midrule
    \multirow{4}{*}{Stage 2}
      & Humanoid Everyday \citep{zhao2026humanoid}  & 16.470  & 4,064   & 1,779,287 \\
      & UniFoLM-WBT \citep{unitree2026unifolmwbt}              & 22.960  & 3,656   & 2,479,519 \\
      & HIW-500  \citep{hiw500_2026}            & 378.150 & 23,743  & 40,839,947 \\
    \cmidrule(l){2-5}
      & \cellcolor[rgb]{ .949,  .949,  .949}\textsc{Total}
      & \cellcolor[rgb]{ .949,  .949,  .949}\textit{417.580}
      & \cellcolor[rgb]{ .949,  .949,  .949}\textit{31,463}
      & \cellcolor[rgb]{ .949,  .949,  .949}\textit{45,098,753} \\
    \midrule
    \multirow{3}{*}{Stage 3}
      & HumanoidArena \citep{DBLP:journals/corr/abs-2606-17833}       & 3.030  & 700     & 545,454 \\
      & Pick-and-Drop-Cola Task (Ego) & 2.456 & 400     & 442,074 \\
      & Pick-and-Drop-Cola Task (Tele-Op)  & 0.826 & 110     & 148,701 \\
    \bottomrule
  \end{tabular}
  \end{center}
\end{table}

\subsubsection{Stage 1: Egocentric Pre-Training}

Egocentric pre-training aims to initialize \model for language-conditioned humanoid action prediction using large-scale egocentric experience.
To this end, we combine human demonstrations from EgoDex \citep{hoque2026egodex}, EgoVerse \citep{DBLP:journals/corr/abs-2604-07607}, and Xperience-10M \citep{xperience10m} with simulated robot demonstrations from BEHAVIOR-1K \citep{li2022behaviork}, expressing heterogeneous motion supervision in the SONIC action space.
Although the BEHAVIOR-1K demonstrations originate from robot simulation, we include these demonstrations into the first training stage, broadly defined here as demonstrations collected without teleoperating real-world robots.
Specifically, we use SONIC motion tokens for EgoDex and construct corresponding motion supervision for EgoVerse from the Aria bimanual subset.
After checking tracking validity, calibration stability, segment duration, and G1 retargeting constraints, we retain 47.08 hours of EgoVerse segments.
We then encode the retained head and wrist trajectories into SONIC motion tokens with a fixed lower-body standing reference.
In comparison, Xperience-10M provides full-body motion references, which we convert into the reference format required by the SONIC encoder.
For BEHAVIOR-1K, a SONIC planner generates G1 motion from upper-body pose, base-motion, and height references.
Alongside these motion tokens, we derive binary hand-control labels from hand keypoints for EgoVerse and Xperience-10M and from gripper commands for BEHAVIOR-1K, while excluding unavailable EgoDex hand-action labels from supervision.
To provide robot-state inputs, we replay the motion tokens on a simulated G1 using the SONIC controller in simulator; for BEHAVIOR-1K, we collect paired motion tokens and states during planner-driven execution.
From these simulation records, we construct a 34-dimensional pseudo-proprioceptive state comprising 29 joint angles, two hand-state entries, and the three-dimensional gravity direction expressed in the body frame.
Within this representation, the hand-state entries use action-label proxies when available and retain simulated hand states for EgoDex.
Finally, we train each VLA variant to predict 50-step action chunks on a 50 Hz control timeline from the corresponding image, task instruction, and normalized pseudo-proprioceptive state, masking padded time steps and unavailable hand-action labels.

\subsubsection{Stage 2: Humanoid Adaptation}

Humanoid adaptation aims to transfer the policy learned from human and simulated experience to real humanoid operation.
Although egocentric pre-training provides diverse motion supervision, the visual observations and pseudo-proprioceptive states can differ from those encountered by a real-world robot.
To address this gap, we continue training each pretrained VLA variant on real G1 demonstrations, organizing heterogeneous data sources into a shared action and state interface.
Specifically, for Humanoid Everyday \citep{zhao2026humanoid}, we select the G1 subset and use SONIC actions and corresponding proprioceptive states.
For UniFoLM-WBT \citep{unitree2026unifolmwbt}, we combine demonstrations collected with different hand types using SONIC motion tokens and standardized binary hand-control labels.
For HIW-500 \citep{hiw500_2026}, we augment motion tokens with hand-control labels reconstructed from controller events, retaining each hand state until the next opening or closing event.
Alongside these labels, we construct proprioceptive inputs from recorded body joint positions, observation-side hand events, and gravity directions derived from the recorded controller pivot orientation.
Using this common representation, we jointly adapt the vision-language backbone and action prediction modules to real robot observations and motion supervision before downstream task fine-tuning.

\subsubsection{Stage 3: Downstream Fine-Tuning}

Downstream fine-tuning adapts the policy to target tasks and environments.
Specifically, we continue training the policy separately for HumanoidArena \citep{DBLP:journals/corr/abs-2606-17833} and the real-world cola pick-and-drop task.
For real-world adaptation, collecting task-specific robot demonstrations through teleoperation is costly and time-consuming.
In particular, joint overheating during data collection can trigger robot shutdowns, interrupting demonstrations and raising safety concerns.
To reduce reliance on robot teleoperation, we supplement robot demonstrations with egocentric human demonstrations collected using a PICO 4 Ultra and trackers attached to the wrists, waist, and ankles.
To incorporate these demonstrations into the shared policy interface, we pair the recordings with SONIC action supervision and obtain pseudo-proprioceptive states through simulated G1 replay.
However, the visible human hands and arms introduce an appearance mismatch with robot observations.
To address this mismatch, we employ the visual editing pipeline of Phantom \citep{lepert2025phantom} to segment and remove human hands and arms, inpaint the exposed regions, and composite rendered robotic hands into the scene.
Within this pipeline, temporally smoothed hand keypoints guide robotic hand rendering, while task-specific grasp constraints and a thumb-visibility prior govern hand configuration and object occlusion in the cola grasping setting.
Throughout this visual transformation, the synthesized videos retain the action targets, pseudo-proprioceptive states, and task instructions associated with the original recordings.
Finally, we combine the synthesized human demonstrations with real robot demonstrations for supervised task adaptation, enabling human task experience to supplement robot data under a visual representation that more closely resembles robot observations.
\section{Experimentation}
\label{sec:experimentation}

We first conduct simulation experiments to iteratively refine the training strategy and egocentric pre-training recipe for humanoid VLAs.
Specifically, experiments on HumanoidArena examine training strategies, the effectiveness of pseudo-proprioceptive states during pre-training, and the scale of egocentric pre-training data.
Building on these investigations, we then explore downstream fine-tuning and task adaptation on the real-world Unitree G1 robot, focusing on the composition of robot and egocentric demonstrations and the visual adaptation of egocentric observations.
Additional implementation details are provided in Appendix~\ref{app:implementation-details}.

\subsection{Simulation Evaluation}
\label{sec:simulation-evaluation}

We evaluate \model on seven HumanoidArena tasks \citep{DBLP:journals/corr/abs-2606-17833}: three Human-Object Interaction (HOI) tasks and four Human-Scene Interaction (HSI) tasks.
Using monocular egocentric RGB observations, language instructions, and robot proprioceptive states, we train a multitask policy for each variant in one run with 100 demonstrations per task.
We then evaluate 60 episodes per task (three seed groups of 20), sharing task configurations across variants.
Finally, we report task success rates, and unweighted task averages for HOI, HSI, and all seven tasks.

\subsubsection{Training Strategy}
\label{sec:training_strategy}

We first investigate which training strategy is most effective for learning humanoid VLAs.
To this end, we compare frozen-backbone training (Frozen), low-rank adaptation (LoRA) \citep{DBLP:conf/iclr/HuSWALWWC22}, and full-parameter fine-tuning (Full) across the three \model variants on HumanoidArena.
As shown in Table~\ref{tab:training-strategy}, LoRA achieves a higher performance than frozen-backbone training for all three variants, supporting the value of backbone adaptation in this setting.
However, full-parameter fine-tuning yields further improvements, achieving average success rates of 55.00\%, 58.33\%, and 60.24\% for \model-OFT, \model-GR00T, and \model-Pi, respectively, and exceeding LoRA by 19.29--38.81 percentage points.
The consistent gains over LoRA suggest that the restricted update space may limit the adaptation of pretrained vision-language representations to humanoid action prediction.
Accordingly, we adopt full-parameter fine-tuning for all subsequent experiments.

\begin{table*}[!htbp]
  \caption{Comparison of training strategies for \model trained without pre-training on HumanoidArena.}
  \label{tab:training-strategy}
  \begin{center}
  \scriptsize
  \resizebox{\textwidth}{!}{%
    \begin{tabular}{@{}lc*{10}{c}@{}}
      \toprule
      \multirow{2}{*}{Model}
        & \multirow{2}{*}{Strategy}
        & \multicolumn{4}{c}{HOI}
        & \multicolumn{5}{c}{HSI}
        & \multirow{2}{*}{Overall AVG ($\uparrow$)} \\
      \cmidrule(lr){3-6}\cmidrule(lr){7-11}
        &
        & Football & DoubleDesk & P\&PBox & AVG ($\uparrow$)
        & OpenDoor & SitSofa & Boxing & VisNavi & AVG ($\uparrow$)
        & \\
      \midrule
      \multirow{3}{*}{\model-OFT}
        & Frozen
        & 0.00 & 0.00 & 0.00 & 0.00
        & 0.00 & 0.00 & 0.00 & 0.00 & 0.00
        & \makebox[3em][r]{0.00} \\
        & LoRA
        & 0.00 & 0.00 & 65.00 & 21.67
        & 0.00 & 8.33 & 38.33 & 1.67 & 12.08
        & \makebox[3em][r]{16.19} \\
        & Full
        & 36.67 & 0.00 & 76.67 & 37.78
        & 88.33 & 53.33 & 80.00 & 50.00 & 67.92
        & \makebox[3em][r]{\textbf{55.00}} \\
      \midrule
      \multirow{3}{*}{\model-GR00T}
        & Frozen
        & 6.67 & 0.00 & 53.33 & 20.00
        & 13.33 & 16.67 & 60.00 & 18.33 & 27.08
        & \makebox[3em][r]{24.05} \\
        & LoRA
        & 16.67 & 0.00 & 61.67 & 26.11
        & 33.33 & 25.00 & 70.00 & 31.67 & 40.00
        & \makebox[3em][r]{34.05} \\
        & Full
        & 40.00 & 8.33 & 85.00 & 44.44
        & 78.33 & 70.00 & 76.67 & 50.00 & 68.75
        & \makebox[3em][r]{\textbf{58.33}} \\
      \midrule
      \multirow{3}{*}{\model-Pi}
        & Frozen
        & 6.67 & 1.67 & 68.33 & 25.56
        & 31.67 & 18.33 & 65.00 & 20.00 & 33.75
        & \makebox[3em][r]{30.24} \\
        & LoRA
        & 16.67 & 3.33 & 70.00 & 30.00
        & 40.00 & 43.33 & 76.67 & 36.67 & 49.17
        & \makebox[3em][r]{40.95} \\
        & Full
        & 50.00 & 8.33 & 81.67 & 46.67
        & 81.67 & 70.00 & 78.33 & 51.67 & 70.42
        & \makebox[3em][r]{\textbf{60.24}} \\
      \bottomrule
    \end{tabular}%
  }
  \end{center}
\end{table*}

\subsubsection{Pseudo Proprioceptive States}
\label{sec:ego-proprio}

Prior studies of robotic manipulation show that VLA policies can rely strongly on the current proprioceptive state when predicting actions \citep{lu2026when}.
However, egocentric human demonstrations lack the corresponding robot-state information.
To assess whether pseudo-proprioceptive states can compensate for this missing input during large-scale egocentric pre-training, we compare the three \model variants with and without pseudo-proprioceptive states during the egocentric pre-training stage.
As shown in Table~\ref{tab:ego-proprio}, incorporating these states improves average success rate by 5.00 and 2.86 percentage points for \model-GR00T and \model-Pi, respectively, suggesting that simulated body configurations provide useful context for learning humanoid action mappings from human demonstrations.
More specifically, SitSofa success increases from 61.67\% to 78.33\% and from 68.33\% to 83.33\%, respectively, consistent with the potential value of body-state information for tasks involving substantial posture changes.
In contrast, \model-OFT decreases from 56.43\% to 54.05\%.
This degradation is associated with the textual state interface.
Specifically, proprioceptive states are serialized as text, and input truncation leaves incomplete state descriptions that introduce ambiguity into action prediction.

\begin{table*}[!htbp]
  \caption{Effect of pseudo-proprioceptive states during egocentric pre-training on HumanoidArena.}
  \label{tab:ego-proprio}
  \begin{center}
  \scriptsize
  \resizebox{\textwidth}{!}{%
    \begin{tabular}{@{}lc*{10}{c}@{}}
      \toprule
      \multirow{2}{*}{Model}
        & \multirow{2}{*}{Ego Proprio}
        & \multicolumn{4}{c}{HOI}
        & \multicolumn{5}{c}{HSI}
        & \multirow{2}{*}{Overall AVG ($\uparrow$)} \\
      \cmidrule(lr){3-6}\cmidrule(lr){7-11}
        &
        & Football & DoubleDesk & P\&PBox & AVG ($\uparrow$)
        & OpenDoor & SitSofa & Boxing & VisNavi & AVG ($\uparrow$)
        & \\
      \midrule
      \multirow{2}{*}{\model-OFT}
        & $\times$
        & 33.33 & 0.00 & 76.67 & 36.67
        & 86.67 & 65.00 & 73.33 & 60.00 & 71.25
        & \makebox[3em][r]{56.43}\hphantom{\,($\downarrow$2.38)} \\
        & $\checkmark$
        & 48.33 & 0.00 & 53.33 & 33.89
        & 75.00 & 55.00 & 73.33 & 73.33 & 69.17
        & \makebox[3em][r]{54.05}\,\red{($\downarrow$2.38)} \\
      \midrule
      \multirow{2}{*}{\model-GR00T}
        & $\times$
        & 63.33 & 3.33 & 76.67 & 47.78
        & 93.33 & 61.67 & 71.67 & 61.67 & 72.09
        & \makebox[3em][r]{61.67}\hphantom{\,($\uparrow$5.00)} \\
        & $\checkmark$
        & 71.67 & 15.00 & 81.67 & 56.11
        & 95.00 & 78.33 & 68.33 & 56.67 & 74.58
        & \makebox[3em][r]{66.67}\,\purple{($\uparrow$5.00)} \\
      \midrule
      \multirow{2}{*}{\model-Pi}
        & $\times$
        & 60.00 & 5.00 & 78.33 & 47.78
        & 90.00 & 68.33 & 73.33 & 60.00 & 72.92
        & \makebox[3em][r]{62.14}\hphantom{\,($\uparrow$2.86)} \\
        & $\checkmark$
        & 63.33 & 11.67 & 76.67 & 50.56
        & 88.33 & 83.33 & 68.33 & 63.33 & 75.83
        & \makebox[3em][r]{65.00}\,\purple{($\uparrow$2.86)} \\
      \bottomrule
    \end{tabular}%
  }
  \end{center}
\end{table*}

\subsubsection{Scaling Egocentric Pre-Training}
\label{sec:scaling-ego}

\noindent
\begin{minipage}[t]{\dimexpr0.5\linewidth-\columnsep\relax}
\vspace{0pt}
We next examine whether scaling egocentric pre-training improves downstream humanoid control.
As shown in Figure~\ref{fig:pretraining-scale}, \model-GR00T and \model-Pi show progressively better downstream performance as egocentric pre-training data increases across the evaluated scales, reaching average performance of 66.67\% and 65.00\%, respectively, with the full corpus.
These gains suggest that additional egocentric experience provides transferable information that benefits downstream control after humanoid adaptation and task-specific fine-tuning.
However, \model-OFT declines at the larger scale, consistent with the previously discussed state truncation issue.
\end{minipage}\hfill
\begin{minipage}[t]{0.5\linewidth}
  \vspace{0pt}
  \centering
  \includegraphics[width=0.63479\linewidth]{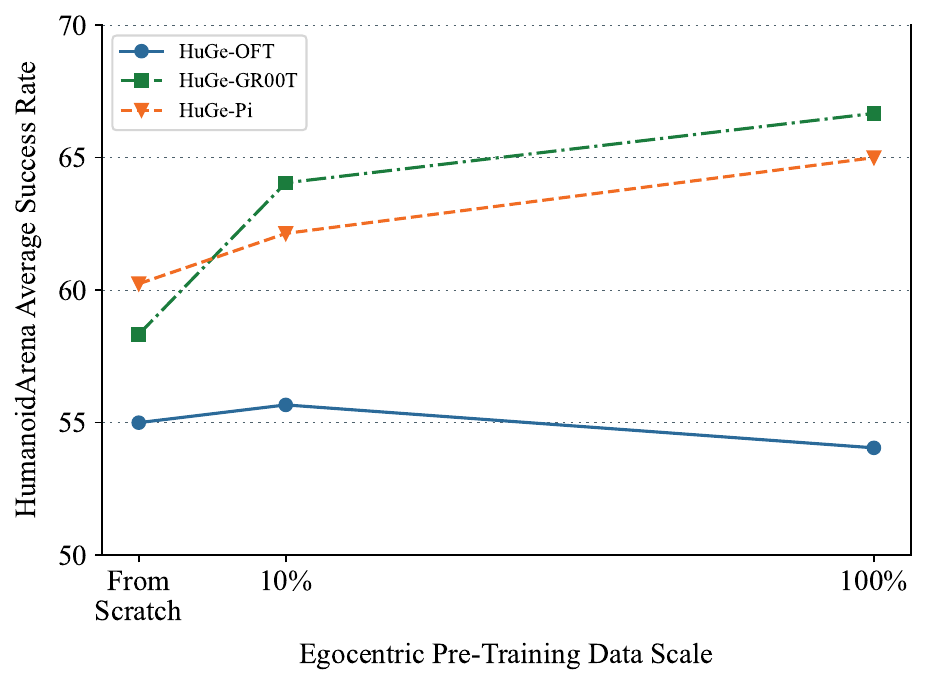}
  \captionsetup{hypcap=false}
  \captionof{figure}{Effect of egocentric pre-training scale.}
  \label{fig:pretraining-scale}
\end{minipage}

\subsubsection{Performance Comparison}
\label{sec:performance-comparison}

To put our results in perspective, we compare the three \model variants to a number of baselines on HumanoidArena.
Specifically, Table~\ref{tab:humanoid-arena-main} includes ACT \citep{DBLP:conf/rss/ZhaoKLF23}, DP \citep{DBLP:conf/rss/ChiFDXCBS23}, FM \citep{DBLP:journals/scirobotics/BarreirosBBCCDFHIIKLLMM26}, and $\pi_{0.5}$ \citep{DBLP:journals/corr/abs-2504-16054} results reported in the original HumanoidArena study \citep{DBLP:journals/corr/abs-2606-17833}.
Additionally, we evaluate GR00T \citep{DBLP:journals/corr/abs-2503-14734} baselines trained from scratch or initialized from GR00T N1.7.
Overall, \model-GR00T and \model-Pi achieve average success rates of 66.67\% and 65.00\%, respectively, exceeding the strongest baseline, $\pi_{0.5}$, by 12.86 and 11.19 percentage points.
Moreover, both variants outperform all listed baselines in the HOI and HSI averages, indicating that the overall gains extend across both interaction categories.
Meanwhile, \model-OFT achieves a lower overall average of 54.05\% but leads on Boxing and VisNavi with 73.33\% success on each task, showing that the strongest overall variant does not dominate every task.
However, all three variants achieve success rates between 0\% and 15.00\% on DoubleDesk.
This limited performance may be related to the prevalence of demonstrations containing repeated grasping attempts and the emphasis on whole-body humanoid data during pre-training.
To address this limitation, future work could incorporate large-scale robotic arm manipulation data into pre-training to improve fine-grained grasping.

\begin{table*}[!htbp]
  \caption{Main results on HumanoidArena. Results marked with $^{\dagger}$ are retrieved from the original HumanoidArena paper. All values are success rates (\%).}
  \label{tab:humanoid-arena-main}
  \begin{center}
  \resizebox{\textwidth}{!}{%
    \begin{tabular}{@{}lc*{10}{c}@{}}
      \toprule
      \multirow{2}{*}{Model}
        & \multirow{2}{*}{Pre-Trained}
        & \multicolumn{4}{c}{HOI}
        & \multicolumn{5}{c}{HSI}
        & \multirow{2}{*}{Overall AVG ($\uparrow$)} \\
      \cmidrule(lr){3-6}\cmidrule(lr){7-11}
        &
        & Football & DoubleDesk & P\&PBox & AVG ($\uparrow$)
        & OpenDoor & SitSofa & Boxing & VisNavi & AVG ($\uparrow$)
        & \\
      \midrule
      ACT \citep{DBLP:conf/rss/ZhaoKLF23}$^{\dagger}$ & \na
        & 20.00 & 30.00 & 60.00 & 36.67
        & 58.30 & 21.70 & 58.30 & 28.30 & 41.65
        & 39.51 \\
      DP \citep{DBLP:conf/rss/ChiFDXCBS23}$^{\dagger}$ & \na
        & 53.30 & 35.00 & 0.00 & 29.43
        & 83.30 & 83.30 & 21.70 & 50.00 & 59.58
        & 46.66 \\
      FM \citep{DBLP:journals/scirobotics/BarreirosBBCCDFHIIKLLMM26}$^{\dagger}$ & \na
        & 21.70 & 20.00 & 0.00 & 13.90
        & 76.70 & 48.30 & 18.30 & 38.30 & 45.40
        & 31.90 \\
      $\pi_{0.5}$ \citep{DBLP:journals/corr/abs-2504-16054}$^{\dagger}$ & \na
        & 10.00 & 30.00 & 75.00 & 38.33
        & 85.00 & 70.00 & 71.70 & 35.00 & 65.43
        & 53.81 \\
      GR00T \citep{DBLP:journals/corr/abs-2503-14734} & \na
        & 0.00 & 0.00 & 75.00 & 25.00
        & 11.67 & 13.33 & 68.33 & 36.67 & 32.50
        & 29.29 \\
      GR00T \citep{DBLP:journals/corr/abs-2503-14734} & GR00T N1.7
        & 13.33 & 0.00 & 76.67 & 30.00
        & 8.33 & 75.00 & 11.67 & 26.67 & 30.42
        & 30.24 \\
      \midrule
      \rowcolor{lightblue}
      \model-OFT & $\checkmark$
        & 48.33 & 0.00 & 53.33 & 33.89
        & 75.00 & 55.00 & 73.33 & 73.33 & 69.17
        & 54.05 \\
      \rowcolor{lightblue}
      \model-GR00T & $\checkmark$
        & 71.67 & 15.00 & 81.67 & 56.11
        & 95.00 & 78.33 & 68.33 & 56.67 & 74.58
        & \textbf{66.67} \\
      \rowcolor{lightblue}
      \model-Pi & $\checkmark$
        & 63.33 & 11.67 & 76.67 & 50.56
        & 88.33 & 83.33 & 68.33 & 63.33 & 75.83
        & 65.00 \\
      \bottomrule
    \end{tabular}%
  }
  \end{center}
\end{table*}

\subsection{Real-World Evaluation}
\label{sec:real-world-evaluation}

\begin{figure}[!t]
  \centering
  \includegraphics[width=\linewidth]{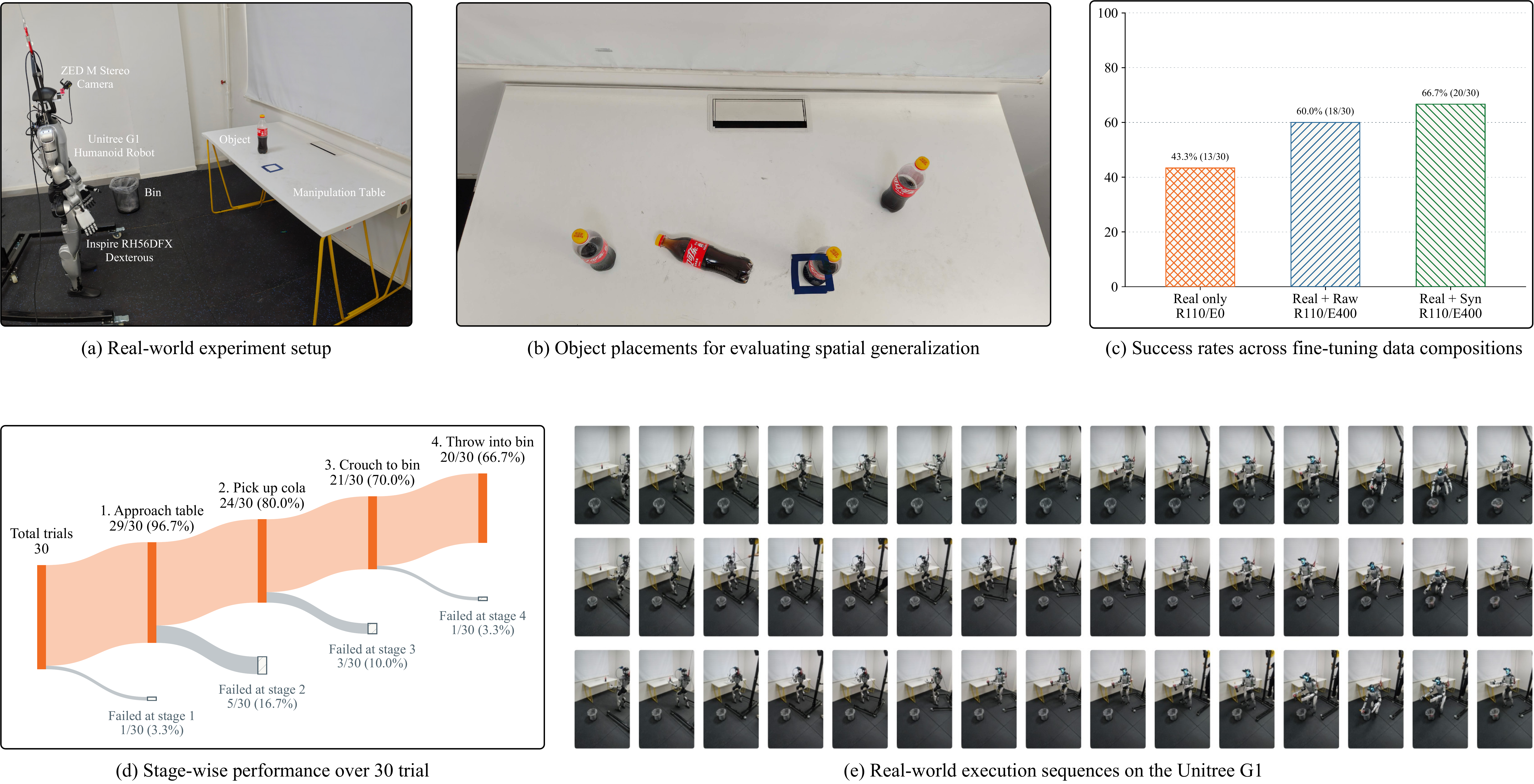}
  \caption{Real-world evaluation of the cola pick-and-drop task on the Unitree G1. In (c), R and E denote real-world robot and egocentric demonstration counts, respectively, while Raw and Syn denote raw and visually synthesized egocentric demonstrations. In (d), flow widths and percentages are relative to all 30 trials, with failures recorded at the first failed stage.}
  \label{fig:real}
\end{figure}

Since real human egocentric demonstrations cannot be collected directly in simulated task environments, we conduct real-world experiments to investigate downstream fine-tuning with egocentric data.
We evaluate \model on a cola pick-and-drop task using a Unitree G1 equipped with Inspire dexterous hands, as shown in Figure~\ref{fig:real}(a).
Specifically, the robot must approach a table, pick up a bottle, turn and walk toward a bin, and crouch to place the bottle inside.
By combining locomotion, grasping, and object placement, this task is designed to assess coordinated whole-body control over a long-horizon sequence.
Furthermore, we randomize the bottle position on the tabletop to assess generalization, as illustrated in Figure~\ref{fig:real}(b).
For these real-world experiments, we use only the $\pi$-style architecture of \model.
To assess the contribution of egocentric data, we compare fine-tuning on 110 robot demonstrations alone with augmentation using 400 raw or visually adapted egocentric demonstrations.
For each configuration, we conduct 30 trials and report the success rate as the fraction of trials that complete the full task.

\subsubsection{Visual Adaptation}

\begin{figure}[!t]
  \centering
  \includegraphics[width=\linewidth]{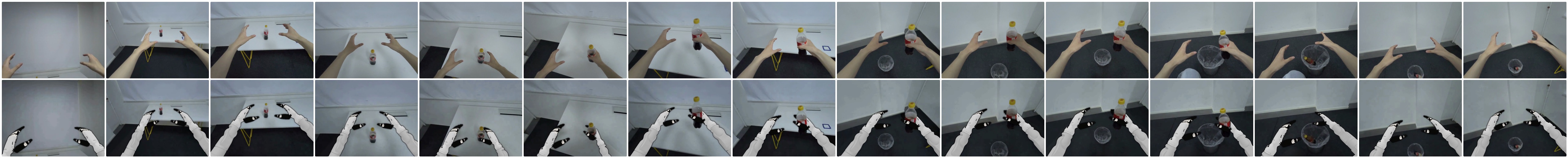}
  \caption{Egocentric observations before (top) and after (bottom) visual adaptation.}
  \label{fig:visual-adapt}
\end{figure}

\begin{figure}[!t]
  \centering
  \includegraphics[width=\linewidth]{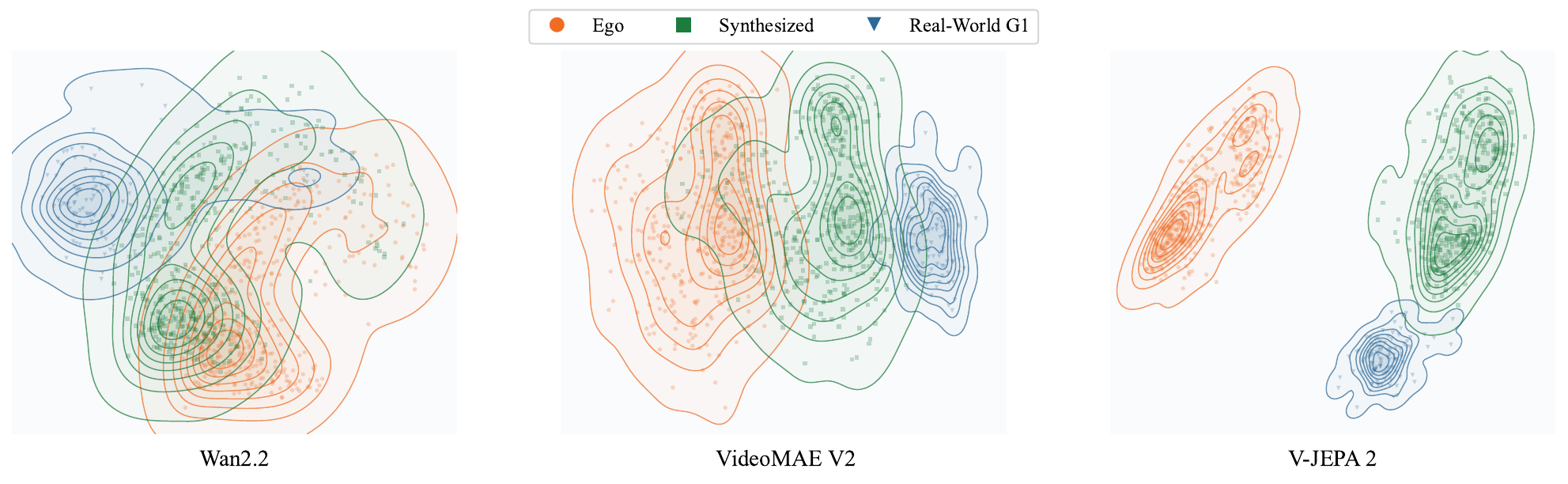}
  \caption{PCA projections of $\ell_2$-normalized features from raw egocentric (Ego), visually adapted (Synthesized), and real-world G1 demonstrations, using Wan2.2 \citep{DBLP:journals/corr/abs-2503-20314}, VideoMAE V2 \citep{DBLP:conf/cvpr/WangHZTHWWQ23}, and V-JEPA 2 \citep{DBLP:journals/corr/abs-2506-09985} from left to right. Contours show Gaussian kernel density estimates.}
  \label{fig:anchor-projection}
\end{figure}

During downstream fine-tuning, we augment egocentric human demonstrations with pseudo-proprioceptive states similar to the pre-training.
Beyond these state inputs, we address a prominent visual discrepancy between human and robot observations caused by differences in hand and arm appearance.
Specifically, we remove the visible human hands and arms, inpaint the exposed regions, and composite simulation-rendered robotic hands into the videos while preserving the original action, state, and language supervision.
Figure~\ref{fig:visual-adapt} illustrates corresponding frames before and after visual adaptation, illustrating the replacement of human hands and arms with rendered robotic counterparts.
To quantify the resulting visual changes, we compare the Fr\'echet distance (FD) from raw and visually adapted egocentric feature distributions to real robot feature distributions.
Specifically, visual adaptation reduces FD from 2.0311 to 1.3499 (33.54\%) for Wan2.2 \citep{DBLP:journals/corr/abs-2503-20314}, from 7.5697 to 5.5204 (27.07\%) for VideoMAE V2 \citep{DBLP:conf/cvpr/WangHZTHWWQ23}, and from 305.5143 to 263.2697 (13.83\%) for V-JEPA 2 \citep{DBLP:journals/corr/abs-2506-09985}.
These reductions indicate that visually adapted egocentric observations more closely resemble real robot observations across all three evaluated feature spaces.
Consistently, Figure~\ref{fig:anchor-projection} shows synthesized features shifting toward the real robot distributions, providing complementary qualitative evidence of reduced visual discrepancy.

\subsubsection{Downstream Fine-Tuning}

Based on this, we evaluate whether egocentric demonstrations and visual adaptation improve downstream fine-tuning on the real-world cola pick-and-drop task.
As shown in Figure~\ref{fig:real}(c), supplementing 110 robot demonstrations with 400 raw egocentric demonstrations increases success from 43.3\% (13/30) to 60.0\% (18/30), supporting the value of human task experience under a fixed robot demonstration budget.
Furthermore, using visually adapted versions of the same egocentric demonstrations raises success to 66.7\% (20/30).
With action, state, and language supervision unchanged, these two additional successful trials suggest that visual adaptation improves the use of human demonstrations for downstream control, consistent with the reduced visual discrepancy observed in the preceding analysis.
At the task level, Figure~\ref{fig:real}(d) reports 20 successful completions across 30 trials, with bottle pickup accounting for five of the ten failed trials and representing the largest source of failure.
During deployment, we observed balance instability during bottle grasping, which we attribute in part to the weight of the dexterous hands.
This instability contributes to unsuccessful grasp attempts and highlights the physical challenge of maintaining balance during manipulation.
Finally, Figure~\ref{fig:real}(e) presents three execution sequences under different task conditions.
Specifically, the first sequence includes a deliberate perturbation absent from the fine-tuning data: the bottle is knocked over during grasping.
Meanwhile, the second and third sequences use different bottle positions on the tabletop.
These demonstrations provide qualitative evidence of adaptability to object disturbances and spatial generalization across the evaluated bottle placements.

\section{Conclusion}
\label{sec:conclusion}

We presented \model, a family of humanoid VLAs trained on egocentric human and robot demonstrations through a three-stage pipeline.
Building on established VLA architectures and a pretrained whole-body controller, \model incorporates pseudo-proprioceptive states and visual adaptation to support human-to-humanoid transfer.
Experiments on HumanoidArena and a real-world G1 pick-and-drop task support the value of egocentric human experience for whole-body policy learning.

\subsection*{AI Use Statement}
The authors wrote and iteratively revised the manuscript, using generative AI tools, primarily GPT 6 Astra, to assist with wording and language polishing throughout the writing process.
Additionally, Claude Opus 4.1, Claude Opus 4.5, Claude Opus 4.7, Claude Opus 4.8, Claude Fable 5, GPT 5.4, GPT 5.5, GPT 5.6 Sol, and GPT 6 Astra assisted with implementing specific functionality according to project requirements via vibe coding defined by the authors.
For schematic illustration, AI assistance was limited to generating the Ego and Tele-op data-collection illustrations in Stage 3 of Figure~\ref{fig:arch}.
Beyond these assistance tasks, generative AI was not used to formulate research questions or hypotheses, design the methodology or experiments, or interpret the results.
The authors take full responsibility for the final manuscript and associated research artifacts, including AI-assisted text, code, and illustrations.

\subsection*{Ethics Statement}
All human demonstrations newly collected for this work were provided by the authors.
For physical safety during real-world robot experiments, a suspension safety frame was used to provide fall protection.
Additionally, we will provide data processing scripts for the existing datasets without directly redistributing the processed data.

\subsection*{Reproducibility Statement}
The main paper and appendices document the experimental setup to support reimplementation and verification of the reported results.
Specifically, Section~\ref{sec:methodology} describes the model architectures, action representations, data sources and processing procedures, and three-stage training pipeline.
Complementing these descriptions, Appendix~\ref{app:implementation-details} provides training hyperparameters, hardware configurations, image preprocessing, and downstream data sampling, while Appendix~\ref{app:proprioceptive-state} specifies the proprioceptive state representation and normalization.
In addition, Section~\ref{sec:experimentation} describes the HumanoidArena and real-world evaluation settings, episode and trial counts, and success-rate computation, with the corresponding task instructions provided in Appendix~\ref{app:task-instructions}.

\printbibliography[title={References}]

\appendix
\section{Implementation Details}
\label{app:implementation-details}

All \model variants use pretrained Qwen3-VL-2B-Instruct \citep{DBLP:journals/corr/abs-2511-21631} as the vision-language backbone.
Under full-parameter fine-tuning, we jointly update the backbone, projection layers, and action modules, while keeping the SONIC controller fixed.
For distributed training, we use DeepSpeed \citep{DBLP:conf/kdd/RasleyRRH20,DBLP:conf/sc/RajbhandariRRH20} on a single node equipped with eight NVIDIA 4090 GPUs, each with 48 GB of memory.
With this setup, Stage 1 and Stage 2 training take approximately 38 and 12 hours, respectively.
Detailed configurations are provided in Table~\ref{tab:hyper-params}.

\begin{table}[!htbp]
    \caption{Training hyper-parameters of \model.}
    \label{tab:hyper-params}
    \small
    \begin{center}
    \resizebox{\textwidth}{!}{
    \begin{tabular}{cccc|cccc}
        \toprule
        Configuration & Stage 1 & Stage 2 & Stage 3 & Configuration & Stage 1 & Stage 2 & Stage 3 \\
        \midrule
        VLM Learning Rate & \multicolumn{3}{c|}{0.00001} & Projector Learning Rate & \multicolumn{3}{c}{0.000025} \\
        Action Head Learning Rate & \multicolumn{3}{c|}{0.0001} & Learning Schedule & \multicolumn{3}{c}{Cosine Decay} \\
        Warm-up Ratio & 5 & 2 & 1 & Numerical Precision & \multicolumn{3}{c}{BFloat16} \\
        Global Batch Size & \multicolumn{3}{c|}{256 / 192 (OFT-Style only)} & Epoch & 1 & 1 & 20k (Step) \\
        Sample Stride & 10 & 5 & 1 & Optimizer & \multicolumn{3}{c}{AdamW \citep{DBLP:conf/iclr/LoshchilovH19}} \\
        \bottomrule
    \end{tabular}
    }
    \end{center}
\end{table}

All training stages share the same image preprocessing pipeline.
Specifically, RGB frames are downsampled when necessary to limit the longer side to 512 pixels while preserving the aspect ratio, padded with black pixels to form a centered square, and resized to $256\times256$ using bicubic interpolation.
During training, we randomly crop a $243\times243$ region, corresponding to approximately 95\% of each spatial dimension, and resize the crop back to $256\times256$ using bicubic interpolation.
Additionally, we apply brightness, contrast, and saturation jitter sequentially, with factors sampled uniformly from $[0.7,1.3]$, $[0.6,1.4]$, and $[0.5,1.5]$, respectively.
At inference, we replace random cropping with center cropping and disable color jitter.
Finally, the resulting RGB images are passed to the Qwen3-VL processor to construct the visual model inputs.

Across the three real-world configurations, the robot demonstrations remain fixed.
For the two configurations augmented with egocentric demonstrations, raw and visually adapted videos share the same action, state, and language supervision.
In both augmented configurations, robot and egocentric sample starts receive equal expected sampling probability.
Additionally, all three configurations share normalization statistics computed from the preceding training data and both downstream robot and egocentric demonstrations.

\section{Proprioceptive State Representation}
\label{app:proprioceptive-state}

\paragraph{State layout.}
Each \model variant conditions action-chunk prediction on a single 34-dimensional G1 state at the starting index $t$:
\begin{equation}
  s_t = \left[(q_t^{\mathrm{body}})^\top,\; h_t^L,\; h_t^R,\; (g_t^{\mathrm{body}})^\top\right]^\top \in \mathbb{R}^{34},
  \label{eq:proprio-state}
\end{equation}
where $q_t^{\mathrm{body}} \in \mathbb{R}^{29}$ contains body joint positions in radians, $h_t^L,h_t^R \in \{0,1\}$ encode left and right hand states (0 for open and 1 for closed), and $g_t^{\mathrm{body}} \in \mathbb{R}^{3}$ is the unit gravity direction in the body frame.
The corresponding dimension order and units are listed in Table~\ref{tab:proprio-state}.

\begin{table}[!htbp]
  \caption{The 34-dimensional G1 proprioceptive state used by \model. Joint fields give angular positions in radians, and hand states use 0 for open and 1 for closed. Gravity fields give the dimensionless $x$, $y$, and $z$ components of the unit gravity direction in the body frame.}
  \label{tab:proprio-state}
  \small
  \begin{center}
  \setlength{\tabcolsep}{2pt}
  \resizebox{\textwidth}{!}{%
  \begin{tabular}{@{}rlll|rlll@{}}
    \toprule
    Index & Field & Meaning & Unit & Index & Field & Meaning & Unit \\
    \midrule
    0 & \texttt{left\_hip\_pitch\_joint} & Left hip pitch & rad & 17 & \texttt{left\_shoulder\_yaw\_joint} & Left shoulder yaw & rad \\
    1 & \texttt{left\_hip\_roll\_joint} & Left hip roll & rad & 18 & \texttt{left\_elbow\_joint} & Left elbow & rad \\
    2 & \texttt{left\_hip\_yaw\_joint} & Left hip yaw & rad & 19 & \texttt{left\_wrist\_roll\_joint} & Left wrist roll & rad \\
    3 & \texttt{left\_knee\_joint} & Left knee & rad & 20 & \texttt{left\_wrist\_pitch\_joint} & Left wrist pitch & rad \\
    4 & \texttt{left\_ankle\_pitch\_joint} & Left ankle pitch & rad & 21 & \texttt{left\_wrist\_yaw\_joint} & Left wrist yaw & rad \\
    5 & \texttt{left\_ankle\_roll\_joint} & Left ankle roll & rad & 22 & \texttt{right\_shoulder\_pitch\_joint} & Right shoulder pitch & rad \\
    6 & \texttt{right\_hip\_pitch\_joint} & Right hip pitch & rad & 23 & \texttt{right\_shoulder\_roll\_joint} & Right shoulder roll & rad \\
    7 & \texttt{right\_hip\_roll\_joint} & Right hip roll & rad & 24 & \texttt{right\_shoulder\_yaw\_joint} & Right shoulder yaw & rad \\
    8 & \texttt{right\_hip\_yaw\_joint} & Right hip yaw & rad & 25 & \texttt{right\_elbow\_joint} & Right elbow & rad \\
    9 & \texttt{right\_knee\_joint} & Right knee & rad & 26 & \texttt{right\_wrist\_roll\_joint} & Right wrist roll & rad \\
    10 & \texttt{right\_ankle\_pitch\_joint} & Right ankle pitch & rad & 27 & \texttt{right\_wrist\_pitch\_joint} & Right wrist pitch & rad \\
    11 & \texttt{right\_ankle\_roll\_joint} & Right ankle roll & rad & 28 & \texttt{right\_wrist\_yaw\_joint} & Right wrist yaw & rad \\
    12 & \texttt{waist\_yaw\_joint} & Waist yaw & rad & 29 & \texttt{left\_hand} & Left hand state & Binary \\
    13 & \texttt{waist\_roll\_joint} & Waist roll & rad & 30 & \texttt{right\_hand} & Right hand state & Binary \\
    14 & \texttt{waist\_pitch\_joint} & Waist pitch & rad & 31 & \texttt{projected\_gravity[0]} & Gravity direction, $x$ & Scalar \\
    15 & \texttt{left\_shoulder\_pitch\_joint} & Left shoulder pitch & rad & 32 & \texttt{projected\_gravity[1]} & Gravity direction, $y$ & Scalar \\
    16 & \texttt{left\_shoulder\_roll\_joint} & Left shoulder roll & rad & 33 & \texttt{projected\_gravity[2]} & Gravity direction, $z$ & Scalar \\
    \bottomrule
  \end{tabular}%
  }
  \end{center}
\end{table}

\paragraph{Normalization.}
We clip and scale each state dimension using the 1st and 99th percentile bounds, $\ell_i$ and $u_i$:
\begin{equation}
  \hat{s}_{t,i} = 2\frac{\operatorname{clip}(s_{t,i},\ell_i,u_i)-\ell_i}{u_i-\ell_i+10^{-8}}-1.
  \label{eq:proprio-normalization}
\end{equation}
These bounds are computed per dimension from pooled training frames across cumulative stages.
Specifically, Stage 1 uses all Stage 1 training data, Stage 2 pools all training data from Stages 1--2, and Stage 3 pools all training data from Stages 1--3.
At inference, we reuse the Stage 3 normalization statistics saved with the checkpoint.

\section{Task Instructions}
\label{app:task-instructions}

Table~\ref{tab:task-instructions} lists the language instructions used for the seven HumanoidArena tasks \citep{DBLP:journals/corr/abs-2606-17833} and the real-world cola pick-and-drop task.

\begin{table}[!htbp]
    \caption{Task instructions for stage 3.}
    \label{tab:task-instructions}
    \small
    \begin{center}
    \resizebox{0.85\textwidth}{!}{
    \begin{tabular}{@{}lll@{}}
      \toprule
      Setting & Task & Task instruction \\
      \midrule
      \multirow{7}{*}{HumanoidArena}
        & Football & Kick the football into the goal. \\
        & DoubleDesk & Put the hammer from the right table into the basket on the left table. \\
        & P\&PBox & Move the box from the table onto the shelf. \\
        & OpenDoor & Open the door. \\
        & SitSofa & Sit on the sofa. \\
        & Boxing & Strike the green markers on the punching bag. \\
        & VisNavi & Avoid obstacles and move to the yellow marked area. \\
      \midrule
      Real world & Cola pick-and-drop & Walk forward, grab the cola and throw into the trash bin \\
      \bottomrule
    \end{tabular}
    }
    \end{center}
\end{table}

\end{document}